\documentclass{article}
\usepackage[preprint]{neurips_2024}
\usepackage[utf8]{inputenc}
\usepackage[T1]{fontenc}
\usepackage{hyperref}
\usepackage{url}
\usepackage{booktabs}
\usepackage{amsfonts}
\usepackage{amsmath}
\usepackage{nicefrac}
\usepackage{microtype}
\usepackage{graphicx}
\usepackage{xcolor}
\usepackage{multirow}
\usepackage{subcaption}

\title{Knowledge Packs: Zero-Token Knowledge Delivery\\via KV Cache Injection}

\author{
  Andrey Pustovit\\
  Independent Researcher\\
  \texttt{andreyandreevgr@gmail.com}
}

\begin{document}

\maketitle
\footnotetext{Code available at \url{https://github.com/cnails/kv-knowledge-packs}}

\begin{abstract}
RAG wastes tokens. We propose \textbf{Knowledge Packs}: pre-computed KV caches that deliver the same knowledge at zero token cost. For causal transformers, the KV cache from a forward pass on text $F$ is identical to what a joint pass on $F \circ q$ would produce---this follows directly from the causal mask. The equivalence is exact but fragile: wrong chat template formatting causes 6--7pp degradation, which we believe explains prior claims of KV outperforming RAG. With correct formatting: zero divergences across 700 questions on Qwen3-8B and Llama-3.1-8B, up to 95\% token savings.

The KV interface also enables behavioral steering that RAG cannot do. Because RoPE rotates keys but leaves values untouched, contrastive deltas on cached values can nudge model behavior while key arithmetic destroys coherence. The effect sits in mid-layer values (33--66\%), independent directions are nearly orthogonal ($\cos \approx 0$) and compose, and both channels---knowledge and steering---run simultaneously at $\alpha{\leq}0.7$ without interference. No training, no weight modification.
\end{abstract}

\section{Introduction}

RAG \citep{lewis2020rag, guu2020realm} solves the knowledge problem by inserting retrieved text into the prompt. One lookup is cheap. But agents that search repeatedly pay a linear tax: 5 searches can consume 700+ tokens on facts alone, eating into the context budget.

We propose \textbf{Knowledge Packs}: pre-computed KV caches that deliver knowledge at zero token cost. In a decoder-only transformer with causal masking, the KV cache from a standalone forward pass on text $F$ is bit-identical to the KV entries for $F$ in a joint pass on $F \circ q$. We call this \textbf{KV--Prefix Equivalence}. Across 500 HotpotQA questions, KV injection and text-in-prompt produce byte-identical outputs.

Same accuracy, $O(1)$ token cost instead of $O(n)$.

The catch is that KV caches must use the model's chat template. Raw text without special tokens is out-of-distribution for instruction-tuned models and degrades accuracy by 6--7pp. We suspect this is why some prior work reports KV outperforming RAG: the KV and RAG conditions were formatted differently.

There is a second use for KV caches. Contrastive deltas applied to cached values (keys untouched) can steer generation style. At $\alpha{\leq}0.5$, this works alongside knowledge delivery without interference.

We make four contributions. First, we prove KV--prefix equivalence and verify it empirically (zero divergences across 700 questions on Qwen3-8B and Llama-3.1-8B), while identifying chat template formatting as a critical pitfall that costs 6--7pp when ignored. Second, we show that KV injection saves up to 95\% of tokens in multi-step accumulation and scales to 5{,}000+ facts via banked routing. Third, we demonstrate that contrastive V-deltas can steer model behavior through KV states that lie outside the image of any token sequence---the effect localizes to mid-layer values and independent directions compose. Fourth, we combine knowledge delivery and value steering in a single cache at $\alpha{\leq}0.7$.

\section{Method}

\subsection{KV Cache Injection}

We pre-compute the KV cache for fact sentences offline and load it as a prefix at inference time.

\paragraph{Write phase (offline, once).} Given fact sentences $\{f_1, \ldots, f_n\}$, we format them using the model's chat template as a system message:

\begin{equation}
    F_{\text{chat}} = \texttt{chat\_template}(\texttt{system}: f_1 \circ \cdots \circ f_n)
\end{equation}

We tokenize $F_{\text{chat}}$ to obtain $\mathbf{x}_F = (x_1, \ldots, x_T)$ and run a forward pass:
\begin{equation}
    \text{KV}_F = \text{Model}(\mathbf{x}_F).\text{past\_key\_values}
\end{equation}
producing $(\mathbf{K}^{(l)}, \mathbf{V}^{(l)}) \in \mathbb{R}^{T \times d_k} \times \mathbb{R}^{T \times d_v}$ for each layer $l$.

\paragraph{Read phase (per query).} Given query $q$, we format it as the user message (without system message, since it is already in KV) and generate:
\begin{equation}
    \text{output} = \text{Model.generate}(\mathbf{x}_q, \; \text{past\_key\_values}=\text{KV}_F, \; \text{attention\_mask}=\mathbf{1}_{T+S})
\end{equation}

\paragraph{Chat template requirement.} Fact text \textit{must} be wrapped in the model's chat template before computing the KV cache. Instruction-tuned models expect special tokens at the start of input (e.g., \texttt{<|im\_start|>system} for Qwen). Without them, 7pp degradation on HotpotQA; on structured generation tasks, 0\% accuracy.

\subsection{KV--Prefix Equivalence Property}

\begin{quote}
\textbf{Property.} For a decoder-only transformer with causal attention, let $F$ be a fact sequence and $q$ a query. Then:
$$\text{KV}(F) \oplus \text{generate}(q \mid \text{KV}(F)) \equiv \text{generate}(F \circ q)$$
That is, pre-computing KV from $F$ alone and generating $q$ with that cache produces identical output to generating from the concatenation $F \circ q$.
\end{quote}

Follows from causal masking: token $x_i$ attends only to $j \leq i$, so KV entries at positions $1, \ldots, T$ are the same with or without query tokens appended. Proof by induction on layers in Appendix~\ref{app:proof}. Does not hold for bidirectional or cross-attention.

\subsection{Banked Routing}

For scaling beyond a single context window, we partition facts into banks via k-means clustering on BGE-large embeddings, with $B = \lceil N/20 \rceil$ banks. At query time: (1) embed query with BGE, (2) select nearest bank centroid, (3) rank facts within bank, (4) recompute KV for top fact on-the-fly (${\sim}$5\,ms). Storage: $<$1\,KB/fact (text + embedding), yielding 4\,MB for 5{,}000 facts.

\subsection{KV Composition}

To compose independently-built caches: process pack A to get $\text{KV}_A$, then process B with $\text{KV}_A$ as prefix so that B's positions continue from $T_A{+}1$. Simply concatenating two caches does not work---both would start at position 0, breaking RoPE.

\section{Experimental Setup}

\paragraph{Models.} Two model families, chosen for their different chat template formats:
\begin{itemize}
    \item \textbf{Qwen3-8B} \citep{qwen3}: Primary model, fp16 on NVIDIA A100 80GB. Instruction-tuned with ChatML-style template (\texttt{<|im\_start|>}/\texttt{<|im\_end|>}).
    \item \textbf{Llama-3.1-8B-Instruct} \citep{dubey2024llama3}: Secondary model, fp16. Uses a different chat template format (\texttt{<|start\_header\_id|>}/\texttt{<|eot\_id|>}) with auto-injected system preamble (``Cutting Knowledge Date'').
\end{itemize}
We chose two models specifically because their templates differ---if our results only held for one template format, that would be a problem.

\paragraph{Benchmarks.}
\begin{itemize}
    \item \textbf{HotpotQA} \citep{yang2018hotpotqa}: Multi-hop QA with 2 gold + 8 distractor paragraphs per question. $N{=}500$ (Qwen) and $N{=}200$ (Llama) from the dev set (75\% bridge + 25\% comparison).
    \item \textbf{HotpotQA Accumulation}: Same question pool with 1--5 sequential retrieval steps, measuring how token cost scales. $N{=}100$ per level per model.
    \item \textbf{Value Steering}: 15 coding tasks scored for defensive patterns (0--9 scale), tested with 30 contrastive code example pairs (``good'': defensive, error-handling vs.\ ``bad'': sloppy, no checks). Both models.
    \item \textbf{Dual-Channel}: 200 HotpotQA bridge questions with simultaneous knowledge KV and formality V-delta, measuring both factual EM and behavioral effect. Qwen3-8B.
\end{itemize}

\paragraph{Methods compared.}
\begin{itemize}
    \item \textbf{Baseline}: No external knowledge.
    \item \textbf{RAG}: Retrieved paragraphs in system message via chat template.
    \item \textbf{KV-chat}: Same paragraphs as KV cache, built with chat template (same system message as RAG). Identical text and formatting; only the delivery mechanism differs.
    \item \textbf{KV-raw}: Same paragraphs as KV cache, built from raw text without chat template. Included to quantify the chat template effect.
\end{itemize}

This is important: RAG and KV-chat use \textit{identical} system messages, character for character. The only variable is whether the text arrives via prompt or via pre-computed cache. Without this control, formatting differences alone can produce 5--10pp gaps, as we show in Section~\ref{sec:chat_template}.

\paragraph{Retrieval.} BGE-large-en-v1.5 \citep{xiao2023bge} top-2 retrieval from all 10 paragraphs.

\paragraph{Evaluation.} Exact match (gold answer substring in prediction) and token-overlap F1.

\section{Results}

\subsection{HotpotQA: KV = RAG at Zero Token Cost}

Table~\ref{tab:hotpotqa} has the main result.

\begin{table}[t]
\centering
\caption{HotpotQA results. KV-chat and RAG use identical system messages with identical chat template formatting; the only difference is the delivery mechanism.}
\label{tab:hotpotqa}
\begin{tabular}{llccccc}
\toprule
\textbf{Model} & \textbf{Method} & \textbf{Overall} & \textbf{Bridge} & \textbf{Comp.} & \textbf{Tok.} & \textbf{Diverg.} \\
\midrule
\multirow{4}{*}{\shortstack[l]{Qwen3-8B\\($N{=}500$)}}
& Baseline & 28.4\% & 17.1\% & 62.4\% & 0 & --- \\
& RAG (BGE top-2) & 65.2\% & 60.5\% & 79.2\% & 284 & --- \\
& \textbf{KV-chat} & \textbf{65.2\%} & \textbf{60.5\%} & \textbf{79.2\%} & \textbf{0} & \textbf{0/500} \\
& KV-raw (no template) & 59.2\% & 54.1\% & 74.4\% & 0 & --- \\
\midrule
\multirow{4}{*}{\shortstack[l]{Llama-3.1-8B\\($N{=}200$)}}
& Baseline & 29.5\% & 18.7\% & 62.0\% & 0 & --- \\
& RAG (BGE top-2) & 61.5\% & 56.7\% & 76.0\% & 290 & --- \\
& \textbf{KV-chat} & \textbf{61.5\%} & \textbf{56.7\%} & \textbf{76.0\%} & \textbf{0} & \textbf{0/200} \\
& KV-raw (no template) & 56.0\% & 49.3\% & 76.0\% & 0 & --- \\
\bottomrule
\end{tabular}
\end{table}

Zero divergences across 700 comparisons. Qwen: 326 both correct, 174 both wrong, 0 disagreements. Llama: 123/77/0. Byte-identical outputs.

KV-raw (no chat template) costs 6.0pp on Qwen, 5.5pp on Llama. The damage concentrates on bridge questions (6.4pp and 7.4pp); comparison questions are barely affected, probably because single-fact lookups are more robust to formatting noise. If the KV and RAG conditions in a study use different formatting, this alone can account for reported accuracy gaps.

\subsection{Accumulation Scaling: Constant vs.\ Linear Token Cost}
\label{sec:accumulation}

Table~\ref{tab:accumulation} shows multi-step retrieval, where an agent accumulates facts over sequential searches.

\begin{table}[t]
\centering
\caption{Accumulation scaling ($N{=}100$ per level per model). As retrieval steps increase, RAG token cost grows linearly while KV remains constant. Accuracy is identical at every level on both models.}
\label{tab:accumulation}
\begin{tabular}{rlccccr}
\toprule
\textbf{Searches} & \textbf{Model} & \textbf{KV EM} & \textbf{RAG EM} & \textbf{KV tok} & \textbf{RAG tok} & \textbf{Savings} \\
\midrule
\multirow{2}{*}{1} & Qwen3-8B & 43.0\% & 43.0\% & 35 & 176 & 141 (80\%) \\
& Llama-3.1-8B & 42.0\% & 42.0\% & 31 & 188 & 157 (84\%) \\
\midrule
\multirow{2}{*}{2} & Qwen3-8B & 59.0\% & 59.0\% & 35 & 299 & 264 (88\%) \\
& Llama-3.1-8B & 59.0\% & 58.0\% & 31 & 305 & 274 (90\%) \\
\midrule
\multirow{2}{*}{3} & Qwen3-8B & 65.0\% & 65.0\% & 35 & 438 & 404 (92\%) \\
& Llama-3.1-8B & 61.0\% & 61.0\% & 31 & 437 & 406 (93\%) \\
\midrule
\multirow{2}{*}{5} & Qwen3-8B & 64.0\% & 64.0\% & 35 & 739 & 704 (95\%) \\
& Llama-3.1-8B & 63.0\% & 63.0\% & 31 & 724 & 693 (96\%) \\
\bottomrule
\end{tabular}
\end{table}

\begin{figure}[t]
\centering
\includegraphics[width=\textwidth]{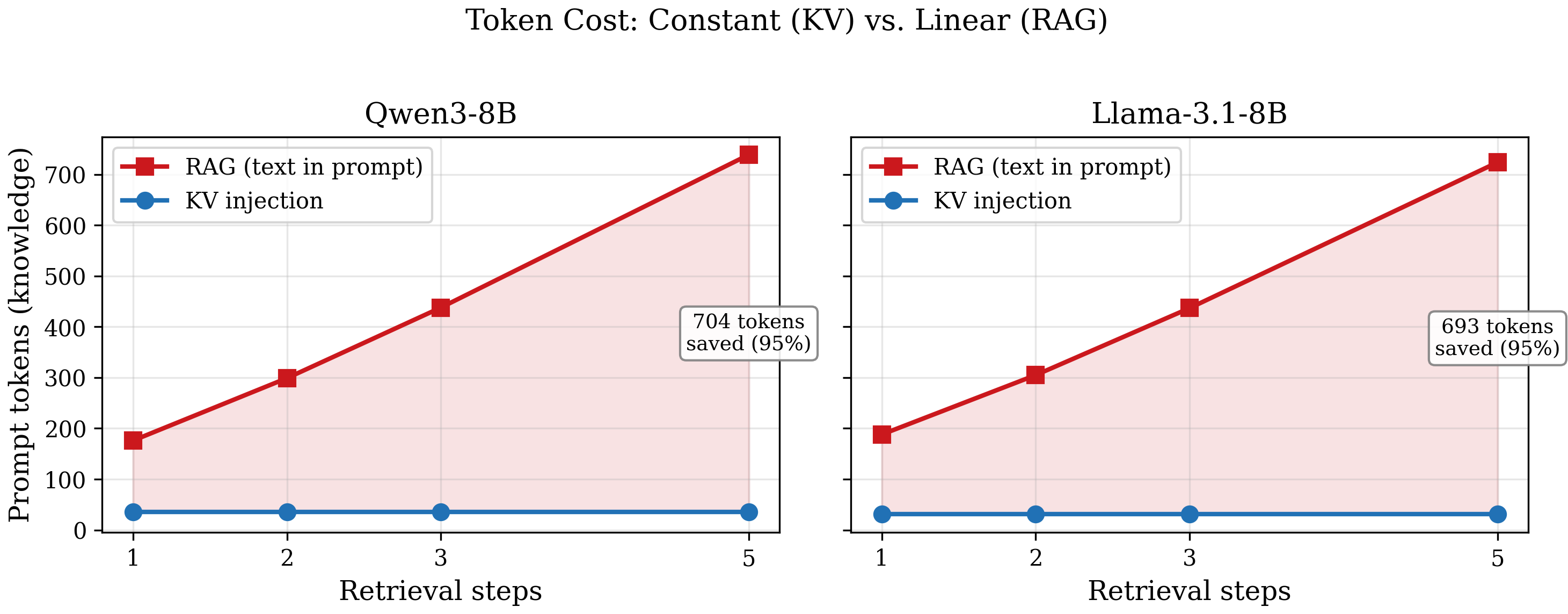}
\caption{Token cost scaling with retrieval steps. RAG prompt cost grows linearly (${\sim}$140--150 tokens per step) while KV injection remains constant (${\sim}$31--35 tokens). Shaded area shows cumulative token savings.}
\label{fig:accumulation}
\end{figure}

Equivalence holds at every level (one 1pp discrepancy at $N{=}100$, noise). Accuracy plateaus around 3 steps.

RAG adds roughly 140--150 tokens per step; KV stays at 31--35 (just the question). At 5 steps this means ${\sim}$700 tokens saved per query. In a 32K context window with average paragraph length of 150 tokens, RAG would hit the context limit at around 200 accumulated facts. KV has no such ceiling.

\subsection{Chat Template Analysis}
\label{sec:chat_template}

Why does some prior work report KV $>$ RAG?

\begin{table}[t]
\centering
\caption{Chat template effect by model and question type.}
\label{tab:chat_template}
\begin{tabular}{llccc}
\toprule
\textbf{Model} & \textbf{Method} & \textbf{Overall} & \textbf{Bridge} & \textbf{Comp.} \\
\midrule
\multirow{3}{*}{Qwen3-8B} & KV-chat & 65.2\% & 60.5\% & 79.2\% \\
& KV-raw & 59.2\% & 54.1\% & 74.4\% \\
& \textbf{Difference} & \textbf{+6.0pp} & \textbf{+6.4pp} & \textbf{+4.8pp} \\
\midrule
\multirow{3}{*}{Llama-3.1-8B} & KV-chat & 61.5\% & 56.7\% & 76.0\% \\
& KV-raw & 56.0\% & 49.3\% & 76.0\% \\
& \textbf{Difference} & \textbf{+5.5pp} & \textbf{+7.4pp} & \textbf{0.0pp} \\
\bottomrule
\end{tabular}
\end{table}

The gap is 5.5--6.0pp overall, concentrated on bridge questions where the model needs to cross-reference multiple facts.

\paragraph{Template split pitfall.} A subtler problem: calling \texttt{apply\_chat\_template} separately for system and user parts can insert duplicate special tokens. On Llama-3.1 this re-adds \texttt{<|begin\_of\_text|>} and an empty system header, which costs another 1.5pp. The fix is to generate the full template (system+user) in one call and split at the boundary. Qwen's template does not have this problem.

These two pitfalls together can shift accuracy by 5--10pp. To be safe, verify that the concatenated KV+prompt token sequence is byte-identical to the single-pass RAG sequence before drawing any conclusions.

\subsection{Scaling with Banked Routing}

We evaluate KV injection at increasing fact counts using synthetic facts with Qwen3-8B (Table~\ref{tab:scaling}).

\begin{table}[t]
\centering
\caption{Scaling KV injection with banked routing (k-means + top-1 selection + lazy recompute).}
\label{tab:scaling}
\begin{tabular}{rrccc}
\toprule
\textbf{Facts} & \textbf{Banks} & \textbf{Routing} & \textbf{Answer} & \textbf{Storage} \\
\midrule
100 & 10 & 100\% & 100\% & 0.4\,MB \\
1{,}000 & 50 & 100\% & 100\% & 1.2\,MB \\
5{,}000 & 250 & 100\% & 100\% & 4.2\,MB \\
\bottomrule
\end{tabular}
\end{table}

Routing accuracy: 100\% at all scales. Storage grows linearly at $<$1\,KB/fact.

\subsection{KV Composition}
\label{sec:composition}

Table~\ref{tab:composition} tests whether independently-built KV caches can be composed.

\begin{table}[t]
\centering
\caption{KV composition on HotpotQA bridge questions (Qwen3-8B, $N{=}200$). Sequential composition preserves positional encoding.}
\label{tab:composition}
\begin{tabular}{lcc}
\toprule
\textbf{Method} & \textbf{Accuracy} & \textbf{Gap vs.\ Single} \\
\midrule
KV Single (one forward pass) & 84.0\% & --- \\
KV Naive (concat, broken positions) & 78.0\% & $-$6.0pp \\
\textbf{KV Sequential (correct positions)} & \textbf{84.5\%} & \textbf{+0.5pp} \\
\bottomrule
\end{tabular}
\end{table}

They can. Naive concatenation loses 6pp (both caches start at RoPE position 0), but sequential composition---running B with $\text{KV}_A$ as prefix---preserves accuracy.

\subsection{Beyond Equivalence: Value-Space Steering}
\label{sec:steering}

Everything above is lossless optimization---same outputs, fewer tokens. But KV caches also give us write access to the model's internal state. What if we write something that no text could produce?

We initially tried arithmetic on both keys and values. Keys broke immediately: coherence dropped to 27\% at $\alpha{=}1.0$. This makes sense once you look at how RoPE \citep{su2024roformer} works---keys are rotated by position ($\mathbf{k}_i = R_{\theta}(i) \cdot W_K \mathbf{x}_i$), and adding rotated vectors from different positions produces garbage. Values are not rotated ($\mathbf{v}_i = W_V \mathbf{x}_i$). They tolerate arithmetic.

So we tried value-only steering. We construct matched pairs of code examples (``good'': defensive, error-handling vs.\ ``bad'': no validation, common pitfalls), keeping structure and variable names identical across pairs. From these we build $\text{KV}_{\text{good}}$ and $\text{KV}_{\text{bad}}$ and steer a fresh base cache:
\begin{equation}
    \mathbf{V}^{(l)}_{\text{steered}} = \mathbf{V}^{(l)}_{\text{base}} + \alpha \cdot (\mathbf{V}^{(l)}_{\text{good}} - \mathbf{V}^{(l)}_{\text{bad}}), \quad \mathbf{K}^{(l)}_{\text{steered}} = \mathbf{K}^{(l)}_{\text{base}}
\end{equation}
Keys are left alone. This is activation steering \citep{turner2023activation, li2024inference} applied to cached values instead of hidden states.

Table~\ref{tab:alpha} sweeps $\alpha$. Score is defensive coding quality (0--9, 15 tasks, pattern-matched); coherence is fraction of syntactically valid outputs.

\begin{table}[t]
\centering
\caption{Value steering alpha sweep with 30 contrastive pairs on two models. Optimal $\alpha$ and coherence cliff differ, but both models show significant improvement over baseline at 100\% coherence.}
\label{tab:alpha}
\begin{tabular}{rcccccc}
\toprule
 & \multicolumn{3}{c}{\textbf{Qwen3-8B} (36L)} & \multicolumn{3}{c}{\textbf{Llama-3.1-8B} (32L)} \\
\cmidrule(lr){2-4} \cmidrule(lr){5-7}
$\alpha$ & Score & Coh. & & Score & Coh. & \\
\midrule
0.0 & 0.60 & 100\% & & 1.47 & 100\% & \\
0.5 & 0.87 & 100\% & & 1.87 & 100\% & \\
1.0 & 1.13 & 100\% & & 2.00 & 100\% & \\
\textbf{1.5} & 1.60 & 100\% & & \textbf{2.47} & \textbf{100\%} & \textbf{+68\%} \\
\textbf{2.0} & \textbf{1.93} & \textbf{100\%} & \textbf{+222\%} & 2.33 & 100\% & \\
3.0 & 0.40 & 20\% & & 1.93 & 100\% & \\
5.0 & 0.00 & 27\% & & 1.33 & 93\% & \\
7.0 & 0.00 & 0\% & & 0.20 & 27\% & \\
\midrule
\multicolumn{4}{l}{\textit{Text-in-prompt}: 3.67} & \multicolumn{3}{l}{\textit{Text-in-prompt}: 3.80} \\
\bottomrule
\end{tabular}
\end{table}

Optimal $\alpha$: 2.0 for Qwen (0.60$\to$1.93), 1.5 for Llama (1.47$\to$2.47). Text-in-prompt scores 3.67 and 3.80 respectively, so this is about half as strong. Qwen is fragile---collapses at $\alpha{=}3.0$ (20\% coherence). Llama is more forgiving, still coherent at 3.0.

Table~\ref{tab:layers} breaks this down by layer.

\begin{table}[t]
\centering
\caption{Layer-selective value steering on two models (each at its optimal $\alpha$, 30 pairs). Mid layers (33--66\%) carry the steering effect on both architectures.}
\label{tab:layers}
\begin{tabular}{lccccc}
\toprule
 & \multicolumn{2}{c}{\textbf{Qwen3-8B} ($\alpha{=}2.0$)} & \multicolumn{2}{c}{\textbf{Llama-3.1-8B} ($\alpha{=}1.5$)} \\
\cmidrule(lr){2-3} \cmidrule(lr){4-5}
\textbf{Layer Range} & $n$ & Score & $n$ & Score \\
\midrule
All (0--100\%) & 36 & 1.93 & 32 & 2.47 \\
Early (0--33\%) & 11 & 1.20 & 10 & 1.33 \\
\textbf{Mid (33--66\%)} & \textbf{12} & \textbf{1.93} & \textbf{11} & \textbf{2.27} \\
Late (66--100\%) & 13 & 0.67 & 11 & 1.47 \\
\midrule
Baseline & --- & 0.60 & --- & 1.47 \\
\bottomrule
\end{tabular}
\end{table}

\begin{figure}[t]
\centering
\includegraphics[width=0.7\textwidth]{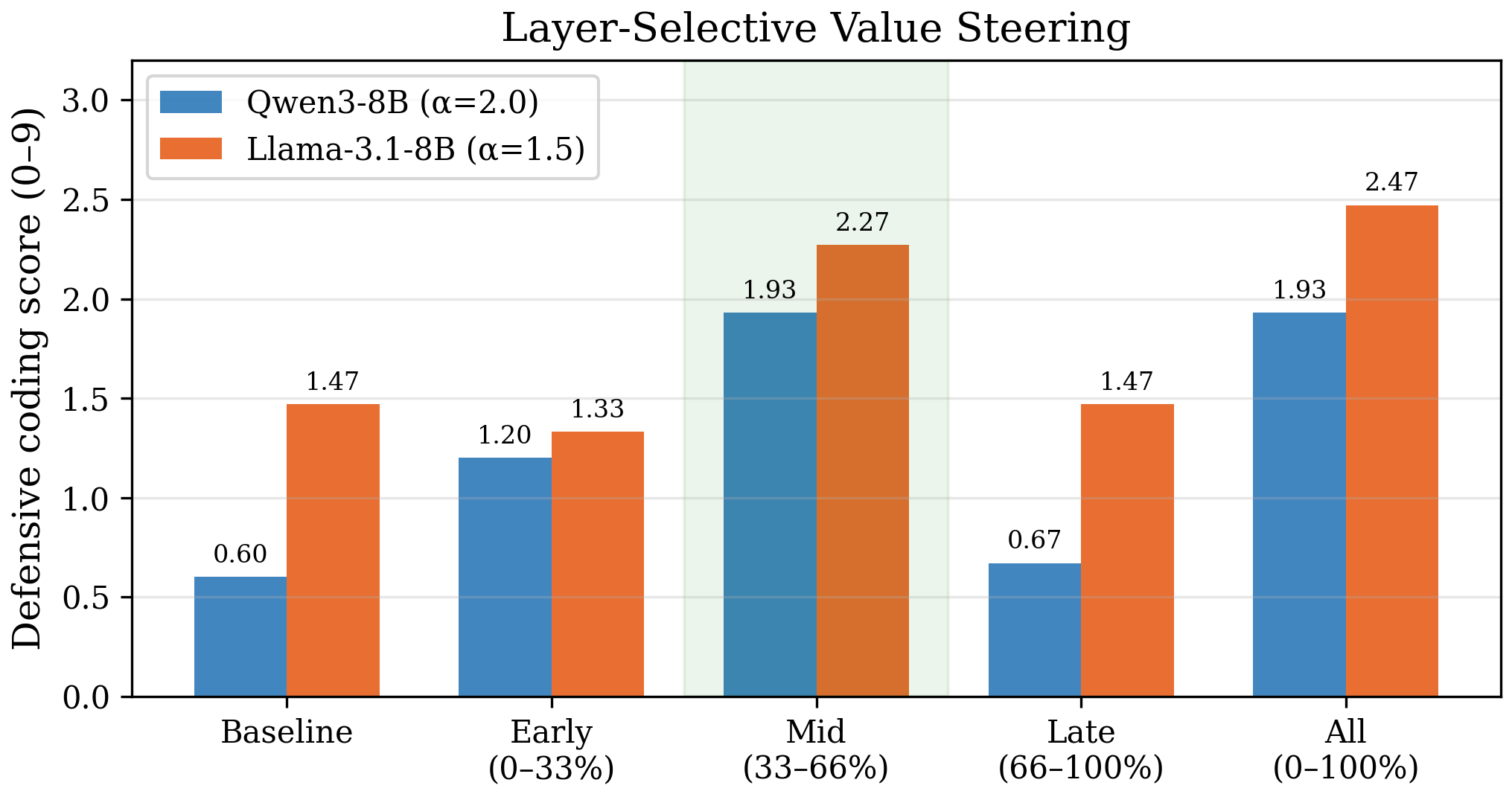}
\caption{Layer-selective value steering on two models. Mid layers (33--66\%, green highlight) capture the full steering effect, matching all-layer steering. Late layers produce scores indistinguishable from baseline.}
\label{fig:layers}
\end{figure}

Mid layers (33--66\%) carry the full effect on Qwen (1.93 = all-layer score) and 92\% on Llama. Late layers do nothing---0.67 on Qwen where baseline is 0.60, and literally 1.47 = 1.47 on Llama.

This is a strong result. It means we can apply steering to only 12 of 36 layers and lose nothing. It also means something about where ``style'' lives in these models, though we will not speculate further here.

5 contrastive pairs and 30 give the same score (1.13 at $\alpha{=}1.0$). The direction saturates fast. Oddly, text-in-prompt gets \textit{worse} with more examples (4.13$\to$3.67).

We also tested composition (Table~\ref{tab:compose}):
\begin{equation}
    \mathbf{V}_{\text{steered}} = \mathbf{V}_{\text{base}} + \sum_{i} \alpha_i \cdot \Delta \mathbf{V}_i, \quad \mathbf{K}_{\text{steered}} = \mathbf{K}_{\text{base}}
\end{equation}

We test two directions: ``defensive coding'' ($\Delta \mathbf{V}_A$: error handling, type checks) and ``documentation'' ($\Delta \mathbf{V}_B$: docstrings, type hints, comments). The cosine similarity between them is $0.0003 \pm 0.012$, i.e., they are nearly orthogonal.

\begin{table}[t]
\centering
\caption{Composable value steering on two models. ``Def.'' counts error-handling patterns (0--9), ``Doc.'' counts docstring/type-hint patterns (0--11). Retention shows A+B score as percentage of single-direction score.}
\label{tab:compose}
\begin{tabular}{lcccccc}
\toprule
 & \multicolumn{3}{c}{\textbf{Qwen3-8B} ($\alpha{=}2.0$)} & \multicolumn{3}{c}{\textbf{Llama-3.1-8B} ($\alpha{=}1.5$)} \\
\cmidrule(lr){2-4} \cmidrule(lr){5-7}
\textbf{Condition} & Def. & Doc. & Coh. & Def. & Doc. & Coh. \\
\midrule
Baseline & 0.40 & 0.00 & 100\% & 1.47 & --- & 100\% \\
\midrule
$A$ only (defensive) & 1.27 & 1.00 & 100\% & 1.87 & 3.80 & 100\% \\
$B$ only (documented) & 0.73 & 3.80 & 100\% & 1.93 & 3.53 & 100\% \\
\textbf{$A{+}B$ composed} & \textbf{1.67} & \textbf{3.40} & \textbf{100\%} & \textbf{2.13} & \textbf{4.07} & \textbf{100\%} \\
\midrule
Def.\ retention & \multicolumn{3}{c}{131\%} & \multicolumn{3}{c}{114\%} \\
Doc.\ retention & \multicolumn{3}{c}{89\%} & \multicolumn{3}{c}{115\%} \\
\midrule
$\cos(\Delta V_A, \Delta V_B)$ & \multicolumn{3}{c}{0.0003} & \multicolumn{3}{c}{0.040} \\
\bottomrule
\end{tabular}
\end{table}

A+B together is \textit{better} than either alone: 131\%/89\% retention on Qwen, 114\%/115\% on Llama. The deltas are nearly orthogonal ($\cos{=}0.0003$ on Qwen, $\cos{=}0.040$ on Llama), which explains why they compose without interference.

$\mathbf{V}_{\text{base}} + \sum_i \alpha_i \cdot \Delta \mathbf{V}_i$ is not in the image of any token sequence---no text produces this KV state. Everything in this subsection (mid-layer localization, composability, near-orthogonality) replicates on both Qwen and Llama.

\subsection{Dual-Channel Architecture: Knowledge + Steering}
\label{sec:dual}

Can we do both at once---knowledge via full KV, steering via mid-layer V-delta? 200 HotpotQA bridge questions, Qwen3-8B, gold paragraphs as knowledge, ``formal/detailed'' V-delta. Six conditions in Table~\ref{tab:dual}.

\begin{table}[t]
\centering
\caption{Dual-channel experiment: knowledge delivery + value steering simultaneously (Qwen3-8B, $N{=}200$ HotpotQA bridge questions). EM = exact match, Form. = formality score, Words = average response length.}
\label{tab:dual}
\begin{tabular}{llccc}
\toprule
& \textbf{Condition} & \textbf{EM} & \textbf{Form.} & \textbf{Words} \\
\midrule
A & No knowledge, no steering & 20.5\% & 0.14 & 16 \\
B & Knowledge only (KV) & 72.5\% & 0.20 & 35 \\
C & Steering only (V-delta) & 15.0\% & 0.32 & 52 \\
D & \textbf{Knowledge + Steering ($\alpha{=}2.0$)} & 40.5\% & 0.32 & 55 \\
E & Text knowledge (RAG) & 73.0\% & 0.13 & 11 \\
F & Text knowledge + text steering & 77.5\% & 0.26 & 48 \\
\bottomrule
\end{tabular}
\end{table}

At $\alpha{=}2.0$ (the steering-only optimum), accuracy drops from 72.5\% to 40.5\%. Too much. Table~\ref{tab:dual_alpha} sweeps lower values.

\begin{table}[t]
\centering
\caption{Alpha sweep for dual mode (knowledge + steering). At $\alpha{\leq}0.7$, factual accuracy is fully preserved while steering effect is measurable. The Pareto frontier is governed by a single scalar.}
\label{tab:dual_alpha}
\begin{tabular}{rccc}
\toprule
$\alpha$ & \textbf{EM} & \textbf{Formality} & \textbf{Words} \\
\midrule
0.0 & 72.5\% & 0.20 & 35 \\
0.1 & 73.0\% & 0.21 & 37 \\
0.2 & 73.0\% & 0.21 & 39 \\
\textbf{0.5} & \textbf{72.0\%} & \textbf{0.23} & \textbf{43} \\
0.7 & 73.0\% & 0.24 & 45 \\
1.0 & 67.0\% & 0.25 & 49 \\
1.5 & 55.0\% & 0.28 & 53 \\
2.0 & 40.5\% & 0.32 & 55 \\
\bottomrule
\end{tabular}
\end{table}

\begin{figure}[t]
\centering
\includegraphics[width=0.65\textwidth]{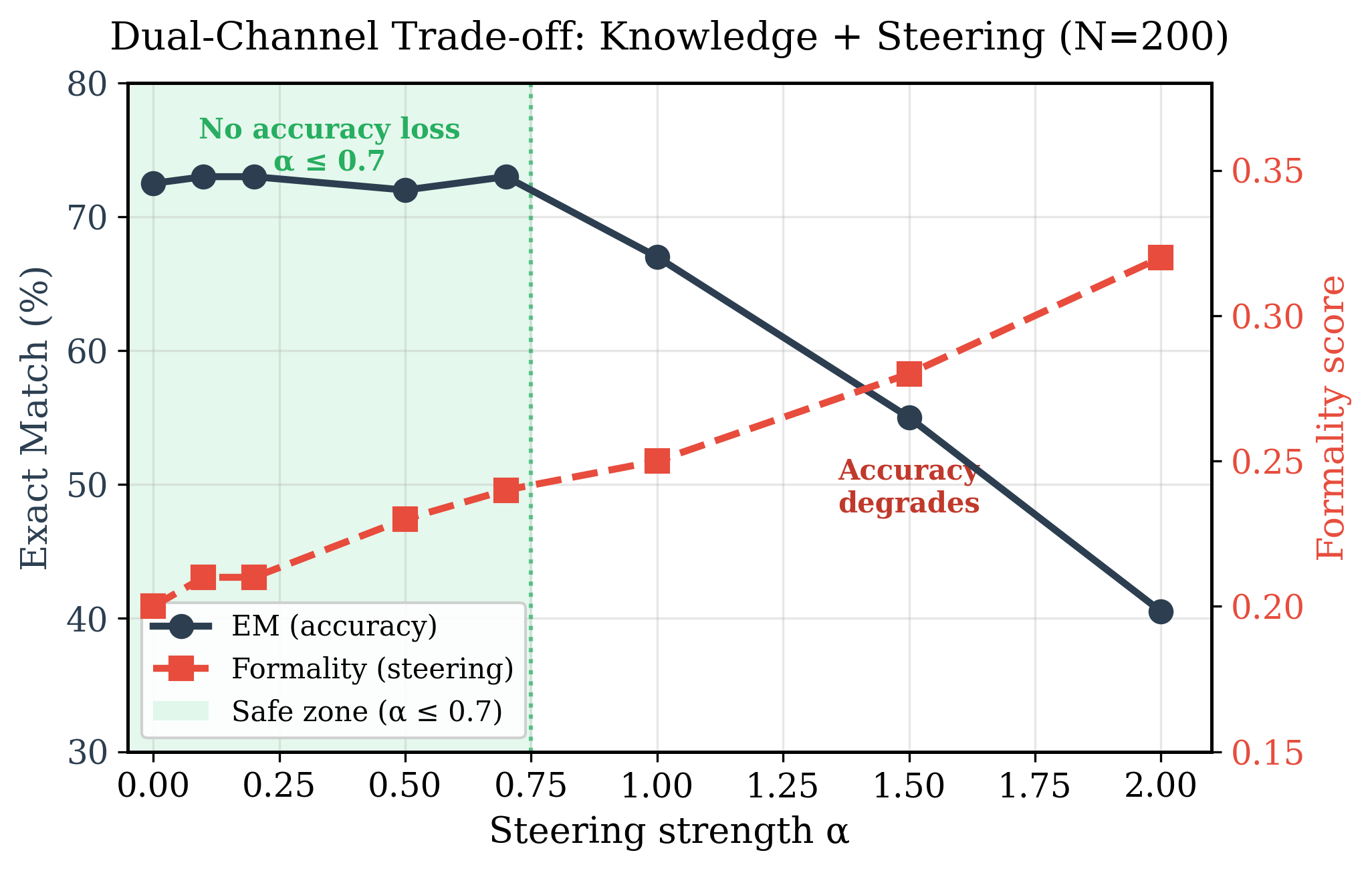}
\caption{Dual-channel trade-off (Qwen3-8B, $N{=}200$). At $\alpha{\leq}0.7$ (green zone), factual accuracy is fully preserved while behavioral steering is measurable. Higher $\alpha$ trades accuracy for stronger steering.}
\label{fig:dual_alpha}
\end{figure}

$\alpha{\leq}0.7$: 72--73\% EM (same as knowledge-only), formality up 20\%, responses 45 vs.\ 35 words. No accuracy cost. At 1.0 accuracy starts to slip (67\%). Past 1.5, steering dominates (55\%).

We think this works because the two channels occupy different parts of the KV space. Knowledge spans all layers, both K and V. Steering only touches mid-layer V. At moderate $\alpha$ they mostly do not overlap.

\section{Analysis and Discussion}

KV = RAG is not a negative result. It is a lossless optimization: swap in a cache, get the same outputs, no tuning. The use cases are pay-per-token APIs, long conversations, agentic workflows. Build cost is ${\sim}$5ms/fact. The main downside is that a Llama cache will not load into Qwen---KV caches are tied to a specific architecture, unlike RAG text.

\paragraph{Chat template as distribution shift.} This one surprised us. Both model families lose 6--7pp with raw-text KV, even though their template formats are completely different (ChatML vs.\ Llama-style headers). Our best explanation is that the special tokens act as mode switches: \texttt{<|im\_start|>system} tells Qwen ``what follows is instructions,'' and without it the model processes the same text differently. The degradation is worst on bridge questions, where the model must cross-reference two passages---single-fact comparison questions are barely affected. We spent a while debugging this before realizing the template was the issue, and we suspect others have hit the same problem without identifying the cause.

This connects to a broader point about prior work claiming KV $>$ RAG. We found three confounds that can produce this: (1)~raw-text KV ($-$6pp), (2)~duplicate BOS from template splitting ($-$1.5pp on Llama), (3)~different system prompt content between conditions. Any one of them can flip the comparison.

\section{Related Work}

\paragraph{RAG.} \citet{lewis2020rag, guu2020realm}, with multi-hop extensions \citep{trivedi2023interleaving, press2023measuring, asai2024selfrag}. KV injection replaces the prompt-insertion step.

\paragraph{KV cache compression and serving.} H2O \citep{zhang2023h2o}, ScissorHands \citep{liu2024scissorhands}, SnapKV \citep{li2024snapkv} prune cache entries; these could be applied on top of our method. Prefix caching in vLLM \citep{kwon2023vllm} and SGLang \citep{zheng2024sglang} shares KV for common prefixes. \citet{shkolnikov2026agent} study persistent KV caches for multi-agent settings. We add the equivalence proof and the chat template requirement, which none of these works address.

\paragraph{Prompt and prefix tuning.} Prefix tuning \citep{li2021prefix} and prompt tuning \citep{lester2021prompt} prepend learned representations but require training. Our caches come from natural language.

\paragraph{Activation steering.} This is the closest line of work to our value steering. Activation addition \citep{turner2023activation} and inference-time intervention \citep{li2024inference} compute steering vectors from contrastive pairs and add them to hidden states during the forward pass. We do the same thing, but on \textit{cached} values rather than live activations, which means the steering is applied once at cache construction time and costs nothing at inference. The key difference is that we cannot touch keys (RoPE makes key arithmetic destructive), so we are limited to the value subspace. Whether this is a fundamental limitation or just a property of RoPE architectures is unclear.

\paragraph{Knowledge editing.} ROME \citep{meng2022rome}, MEMIT \citep{meng2023memit}. Weight editing. We do not modify weights.

\section{Conclusion}

Pre-computed KV caches are a lossless, zero-token replacement for RAG prompt insertion. Zero divergences on 700 questions, two models, up to 95\% token savings at 5 retrieval steps. The main practical requirement is correct chat template formatting.

Separately, the RoPE K/V asymmetry opens up value-space steering: contrastive V-deltas applied to mid layers can nudge model behavior without re-running inference, and independent directions compose. Both channels---knowledge and steering---run simultaneously at $\alpha{\leq}0.7$ without degrading either.

\paragraph{Limitations.} The most obvious limitation is that KV caches do not transfer across models. Value steering, while consistent, only reaches about half the effect of text-in-prompt (1.93--2.47 vs.\ 3.67--3.80 on our coding tasks). We also have limited scale: 15 coding tasks for steering, 200 questions for dual-channel, and steering was only tested on coding style---we do not know if it transfers to, say, tone or factuality. The routing evaluation used synthetic facts that are easy to distinguish; real knowledge bases would be harder. Finally, $\alpha$ must be reduced in dual-channel mode compared to steering-only, which tells us the two subspaces are not fully orthogonal despite what the cosine similarities suggest.


\bibliographystyle{plainnat}
\bibliography{references}

@inproceedings{lewis2020rag,
  title={Retrieval-Augmented Generation for Knowledge-Intensive NLP Tasks},
  author={Lewis, Patrick and Perez, Ethan and Piktus, Aleksandra and Petroni, Fabio and Karpukhin, Vladimir and Goyal, Naman and K{\"u}ttler, Heinrich and Lewis, Mike and Yih, Wen-tau and Rockt{\"a}schel, Tim and others},
  booktitle={Advances in Neural Information Processing Systems},
  volume={33},
  pages={9459--9474},
  year={2020}
}

@inproceedings{guu2020realm,
  title={REALM: Retrieval-Augmented Language Model Pre-Training},
  author={Guu, Kelvin and Lee, Kenton and Tung, Zora and Pasupat, Panupong and Chang, Mingwei},
  booktitle={International Conference on Machine Learning},
  pages={3929--3938},
  year={2020}
}

@inproceedings{yang2018hotpotqa,
  title={HotpotQA: A Dataset for Diverse, Explainable Multi-hop Question Answering},
  author={Yang, Zhilin and Qi, Peng and Zhang, Saizheng and Bengio, Yoshua and Cohen, William W and Salakhutdinov, Ruslan and Manning, Christopher D},
  booktitle={Empirical Methods in Natural Language Processing},
  year={2018}
}

@article{dubey2024llama3,
  title={The Llama 3 Herd of Models},
  author={Dubey, Abhimanyu and others},
  journal={arXiv preprint arXiv:2407.21783},
  year={2024}
}

@article{xiao2023bge,
  title={C-Pack: Packaged Resources To Advance General Chinese Embedding},
  author={Xiao, Shitao and Liu, Zheng and Zhang, Peitian and Muennighoff, Niklas},
  journal={arXiv preprint arXiv:2309.07597},
  year={2023}
}

@inproceedings{zhang2023h2o,
  title={H2O: Heavy-Hitter Oracle for Efficient Generative Inference of Large Language Models},
  author={Zhang, Zhenyu and Sheng, Ying and Zhou, Tianyi and Chen, Tianlong and Zheng, Lianmin and Cai, Ruisi and Song, Zhao and Tian, Yuandong and R{\'e}, Christopher and Barrett, Clark and others},
  booktitle={Advances in Neural Information Processing Systems},
  year={2023}
}

@inproceedings{liu2024scissorhands,
  title={Scissorhands: Exploiting the Persistence of Importance Hypothesis for LLM KV Cache Compression at Test Time},
  author={Liu, Zichang and Desai, Aditya and Liao, Fangshuo and Wang, Weitao and Xie, Victor and Xu, Zhaozhuo and Kyrillidis, Anastasios and Shrivastava, Anshumali},
  booktitle={Advances in Neural Information Processing Systems},
  year={2024}
}

@article{li2024snapkv,
  title={SnapKV: LLM Knows What You are Looking for Before Generation},
  author={Li, Yuhong and Huang, Yingbing and Yang, Bowen and Venkitesh, Bharat and Locatelli, Acyr and Ye, Hanchen and Cai, Tianle and Lewis, Patrick and Chen, Deming},
  journal={arXiv preprint arXiv:2404.14469},
  year={2024}
}

@inproceedings{kwon2023vllm,
  title={Efficient Memory Management for Large Language Model Serving with PagedAttention},
  author={Kwon, Woosuk and Li, Zhuohan and Zhuang, Siyuan and Sheng, Ying and Zheng, Lianmin and Yu, Cody Hao and Gonzalez, Joseph and Zhang, Hao and Stoica, Ion},
  booktitle={Symposium on Operating Systems Principles},
  year={2023}
}

@article{zheng2024sglang,
  title={SGLang: Efficient Execution of Structured Language Model Programs},
  author={Zheng, Lianmin and Yin, Liangsheng and Xie, Zhiqiang and Cheng, Shuo and Huang, Jeff and Zhuang, Siyuan and Shi, Yinmin and Stoica, Ion},
  journal={arXiv preprint arXiv:2312.07104},
  year={2024}
}

@inproceedings{li2021prefix,
  title={Prefix-Tuning: Optimizing Continuous Prompts for Generation},
  author={Li, Xiang Lisa and Liang, Percy},
  booktitle={Association for Computational Linguistics},
  year={2021}
}

@inproceedings{lester2021prompt,
  title={The Power of Scale for Parameter-Efficient Prompt Tuning},
  author={Lester, Brian and Al-Rfou, Rami and Constant, Noah},
  booktitle={Empirical Methods in Natural Language Processing},
  year={2021}
}

@inproceedings{meng2022rome,
  title={Locating and Editing Factual Associations in GPT},
  author={Meng, Kevin and Bau, David and Andonian, Alex and Belinkov, Yonatan},
  booktitle={Advances in Neural Information Processing Systems},
  year={2022}
}

@inproceedings{meng2023memit,
  title={Mass-Editing Memory in a Transformer},
  author={Meng, Kevin and Sharma, Arnab Sen and Andonian, Alex and Belinkov, Yonatan and Bau, David},
  booktitle={International Conference on Learning Representations},
  year={2023}
}

@inproceedings{trivedi2023interleaving,
  title={Interleaving Retrieval with Chain-of-Thought Reasoning for Knowledge-Intensive Multi-Step Questions},
  author={Trivedi, Harsh and Balasubramanian, Niranjan and Khot, Tushar and Sabharwal, Ashish},
  booktitle={Association for Computational Linguistics},
  year={2023}
}

@article{press2023measuring,
  title={Measuring and Narrowing the Compositionality Gap in Language Models},
  author={Press, Ofir and Zhang, Muru and Min, Sewon and Schmidt, Ludwig and Smith, Noah A and Lewis, Mike},
  journal={Findings of EMNLP},
  year={2023}
}

@article{shkolnikov2026agent,
  title={Agent Memory Below the Prompt: Persistent Q4 KV Cache},
  author={Shkolnikov, Mikhail},
  journal={arXiv preprint arXiv:2603.04428},
  year={2026}
}

@article{qwen3,
  title={Qwen3 Technical Report},
  author={{Qwen Team}},
  journal={arXiv preprint arXiv:2505.09388},
  year={2025}
}

@inproceedings{asai2024selfrag,
  title={Self-RAG: Learning to Retrieve, Generate, and Critique through Self-Reflection},
  author={Asai, Akari and Wu, Zeqiu and Wang, Yizhong and Sil, Avirup and Hajishirzi, Hannaneh},
  booktitle={International Conference on Learning Representations},
  year={2024}
}

@article{turner2023activation,
  title={Activation Addition: Steering Language Models Without Optimization},
  author={Turner, Alexander Matt and Thiergart, Lisa and Udell, David and Leech, Gavin and Mini, Ulisse and Pelrine, Monte},
  journal={arXiv preprint arXiv:2308.10248},
  year={2023}
}

@inproceedings{li2024inference,
  title={Inference-Time Intervention: Eliciting Truthful Answers from a Language Model},
  author={Li, Kenneth and Patel, Oam and Vi{\'e}gas, Fernanda and Pfister, Hanspeter and Wattenberg, Martin},
  booktitle={Advances in Neural Information Processing Systems},
  year={2024}
}

@article{su2024roformer,
  title={RoFormer: Enhanced Transformer with Rotary Position Embedding},
  author={Su, Jianlin and Ahmed, Murtadha and Lu, Yu and Pan, Shengfeng and Bo, Wen and Liu, Yunfeng},
  journal={Neurocomputing},
  volume={568},
  year={2024}
}

\appendix
\section{Proof of KV--Prefix Equivalence}
\label{app:proof}

\textbf{Claim.} $\text{KV}^{(l)}_F = \text{KV}^{(l)}_{F \circ q}[1{:}T]$ for all layers $l$.

The argument is simple. Causal mask: position $i$ sees only $j \leq i$. So appending $q$ after position $T$ cannot affect KV entries at positions $1, \ldots, T$. This holds at layer 1 (embeddings depend only on token identity and position), and propagates by induction---layer $l$ at position $i \leq T$ depends only on layer $l{-}1$ outputs at positions $\leq i$, all of which are identical by hypothesis. Given identical caches, generation from $T{+}1$ is deterministic and identical. $\square$

Requires causal masking, position encodings that depend on index only, deterministic decoding.

\section{Synthetic Scaling Tests}
\label{app:scaling}

Scaling evaluation with Qwen3-8B using fictional facts:
\begin{itemize}
    \item \textbf{Single KV cache}: 100\% accuracy at 1{,}000 facts (8{,}434 tokens).
    \item \textbf{Banked KV}: 100\% routing and answer accuracy at 5{,}000 facts with 250 banks.
    \item \textbf{Lazy recompute}: 4.2\,MB storage, 6\,ms per-query overhead.
\end{itemize}

\end{document}